\documentclass[10pt,twocolumn,letterpaper]{article}

\usepackage{cvpr}
\usepackage{times}
\usepackage{epsfig}
\usepackage{graphicx}
\usepackage{amsmath}
\usepackage{amssymb}
\usepackage{booktabs} 
\usepackage{textcomp}
\usepackage{stackengine}
\usepackage{indentfirst}
\usepackage{resizegather}
\usepackage{subcaption}
\usepackage[final]{pdfpages}

\usepackage[pagebackref=true,breaklinks=true,letterpaper=true,colorlinks,bookmarks=false]{hyperref}

\cvprfinalcopy 

\def\cvprPaperID{3867} 

\ifcvprfinal\fi
\begin{document}

\title{Cross-View Image Synthesis using Conditional GANs} 

\author{Krishna Regmi
and
Ali Borji\\
Center for Research in Computer Vision, University of Central Florida\\
{\tt\small krishna.regmi7@gmail.com, aborji@crcv.ucf.edu}
}

\maketitle


\begin{abstract}
\vspace{-10pt}
Learning to generate natural scenes has always been a challenging task in computer vision. It is even more painstaking when the generation is conditioned on images with drastically different views. This is mainly because understanding, corresponding, and transforming appearance and semantic information across the views is not trivial. In this paper, we attempt to solve the novel problem of cross-view image synthesis, aerial to street-view and vice versa, using conditional generative adversarial networks (cGAN). Two new architectures called Crossview Fork (X-Fork) and Crossview Sequential (X-Seq) are proposed to generate scenes with resolutions of 64$\times$64 and 256$\times$256 pixels. X-Fork architecture has a single discriminator and a single generator. The generator hallucinates both the image and its semantic segmentation in the target view. X-Seq architecture utilizes two cGANs. The first one generates the target image which is subsequently fed to the second cGAN for generating its corresponding semantic segmentation map. The feedback from the second cGAN helps the first cGAN generate sharper images. Both of our proposed architectures learn to generate natural images as well as their semantic segmentation maps. The proposed methods show that they are able to capture and maintain the true semantics of objects in source and target views better than the traditional image-to-image translation method which considers only the visual appearance of the scene. Extensive qualitative and quantitative evaluations support the effectiveness of our frameworks, compared to two state of the art methods, for natural scene generation across drastically different views.
\end{abstract}

\vspace{-15pt}

\section{Introduction} 
\vspace{-5pt}
In this work, we address the problem of synthesizing ground-level images from overhead imagery and vice versa using conditional Generative Adversarial Networks~\cite{DBLP:journals/corr/MirzaO14}. Primarily, such models try to generate new images from conditioning variables as input. Preliminary works in GANs utilize unsupervised learning to generate samples from latent representations or from a random noise vector~\cite{goodfellow2014generative}.

\begin{figure}
\centering
\includegraphics[width=0.45\textwidth]{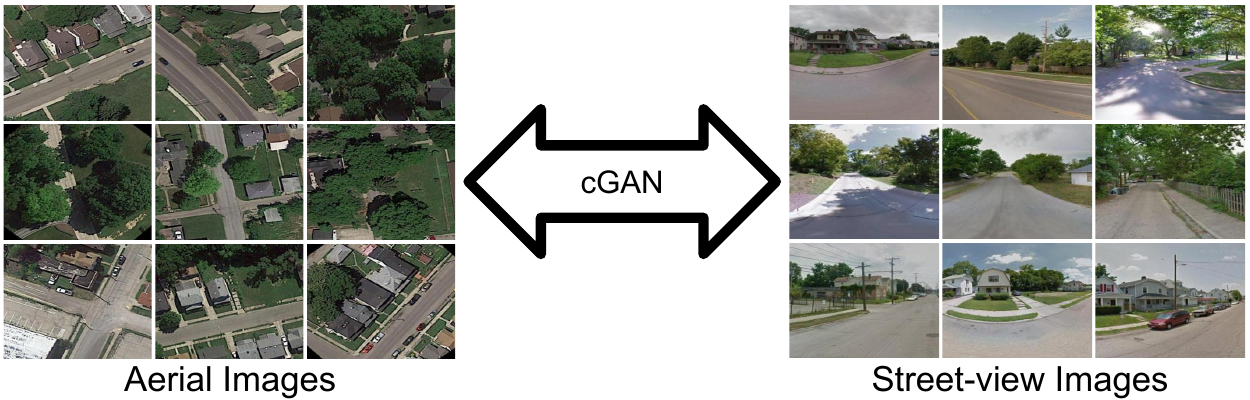}
\vspace{-8pt}
\caption{\small \label{fig:obj} Example images in overhead/aerial view (left) and ground-level/street-view (right). The images reflect the great diversity and richness of features in two views implying that the network needs to learn a lot for meaningful cross-view generation. We propose to use cGANs to solve this problem.}
\vspace{-15pt}
\end{figure}

View synthesis is a long-standing problem in computer vision. This task is more challenging when views are drastically different, fields of views have little or no overlap, and objects are occluded. Furthermore, two objects that are similar in one view may look quite different in another (i.e., the view-invariance problem). For example, the aerial view of a building (i.e., the roof) tells very little about the color and design of the building seen from the street-view. The generation process is generally easier when the image contains a single object in a uniform background. In contrast, when the scene contains multiple objects, generating other view becomes much more challenging. This is due to the increase in underlying parameters that contribute to the variations (e.g., occlusions, shadows, etc). An example scenario, addressed here, is generating street-view (a.k.a ground level) image of a location from its aerial (a.k.a overhead) imagery. Figure \ref{fig:obj} illustrates some corresponding images in two views. 

Isola \textit{et al.} \cite{pix2pix2017} put forward a general-purpose framework to solve multiple image translation tasks. Their work translates images of objects or scenes which are represented by RGB images, gradient fields, edge maps, aerial images, sketches, etcetera between these representations. Thus, their method operates on representations in a single view. Formulating our problem as an image translation task between the views, we use their method as a baseline and extend it for cross-view image generation. 

Inspired by recent works of \cite{pix2pix2017,zhai2017crossview}, we formulate the cross-view image synthesis problem as an image-to-image translation problem and solve it using the conditional generative adversarial network. Previous works in view synthesis \cite{DTB17,10.1007/978-3-319-46493-0_18,10.1007/978-3-319-46478-7_20} have generated images with single objects in them or natural scenes with very little variation in viewing angles between the input and target images. The network learns to copy large parts of image content from the input. Images in each view of our work contain high degree of details and clutter (e.g., trees, cars, roads, buildings, etcetra) along with variations between the corresponding image pairs. Thus, the network needs to learn that the corresponding images in each view need to contain all details and place them in correct positions with proper orientations and inclinations. Perhaps the closest work to ours is the one by Zhai \textit{et al.} \cite{zhai2017crossview}. They generate ground-level panorama from aerial imagery by predicting ground image features from aerial image features and use them to synthesize images. 

In general, some of the challenges pertaining to cross-view synthesis task are as follows. First, aerial images cover wider regions of the ground than the street-view images whereas street-view images contain more details about objects (e.g., house, road, trees) than aerial images. So, not only the information in aerial images is too noisy, but also less informative for street-view image synthesis. Similarly, a network needs to estimate a lot regions to synthesize aerial images. Second, transient objects like cars (also people) are not present at the corresponding locations in image pairs since they are taken at different times. Third, houses that are different in street-view look similar from aerial view. This causes synthesized street-view images to contain buildings with similar color or texture, prohibiting diversity in generated buildings. 
Fourth challenge regards variation among roads in two views due to perspective and occlusions. While the road edges are nearly linear and visible in street-view, they are often occluded by dense vegetations and contorted in aerial view. Fifth, when using model generated segmentation maps as ground truth to improve the quality of generated images, as done here, label noise and model errors introduce some artifacts in the results.




To address the above challenges, we propose the following methods. We start with a simple image-to-image translation network of \cite{pix2pix2017} as a baseline. We then propose two new cGAN architectures that generate images as well as segmentation maps in target view. Addition of semantic segmentation generation to the architectures helps improve the generation of images. The first architecture, called X-Fork, is a slight modification of the baseline, forking at the penultimate block to generate two outputs, target view image and segmentation map. The second architecture, called X-Seq, has a sequence of two baseline networks connected. The target view image generated by the first network is fed to the second network to generate its corresponding segmentation map. Once trained, both architectures are able to generate better images than the baseline that learns to generate the images only. This implies that learning to generate segmentation map along with the image indeed improves the quality of generated image.


\section{Related Works}
\vspace{-5pt}
\subsection{Relating Aerial and Ground-level Images}
\vspace{-5pt}
Zhai \textit{et al.} \cite{zhai2017crossview} explored to predict the semantic layout of ground image from its corresponding aerial image. They used the predicted layout to synthesize ground-level panorama. Prior works relating the aerial and ground imageries have addressed problems such as cross-view co-localization \cite{Lin_2013_CVPR, Vo2016}, ground-to-aerial geo-localization \cite{DBLP:conf/cvpr/LinCBH15} and geo-tagging the cross-view images \cite{workman2015wide}. 

Cross-view relations have also been studied between egocentric (first person) and exocentric (surveillance or third-person) domains for different purposes. Human re-identification by matching viewers in top-view and egocentric cameras have been tackled by establishing the correspondences between the views in \cite{DBLP:conf/eccv/ArdeshirB16}. Soran \textit{et al.} \cite{DBLP:conf/accv/SoranFS14} utilize the information from one egocentric camera and multiple exocentric cameras to solve the action recognition task. Ardeshir \textit{et al.} \cite{DBLP:journals/corr/ArdeshirRB16} learn motion features of actions performed in ego- and exocentric domains to transfer motion information across the two domains. 

\vspace{-5pt}
\subsection{Learning View Transformations}
\vspace{-5pt}
Existing works on viewpoint transformation have been conducted to synthesize novel views of the same objects \cite{DTB17,10.1007/978-3-319-46478-7_20,10.1007/978-3-319-46493-0_18}. Zhou \textit{et al.} \cite{10.1007/978-3-319-46493-0_18} proposed models that learn to copy the pixel information from input view and utilize them to preserve the identity and structure of the objects to generate new views. Tatarchenko \textit{et al.} \cite{10.1007/978-3-319-46478-7_20} trained an encode-decoder network to obtain 3D representation models of cars and chairs which they later used to generate different views of an unseen car or chair image. Dosovitskiy \textit{et al.} \cite{DTB17} learned generative models by training on 3D renderings of cars, chairs and tables and synthesized intermediate views and objects by interpolating between views and models.

\vspace{-5pt}
\subsection{GAN and cGAN}
\vspace{-5pt}
Goodfellow \textit{et al.} \cite{goodfellow2014generative} are the pioneers of Generative Adversarial Networks that is very successful at generating sharp and unblurred images, much better compared to existing methods such as Restricted Boltzmann Machines \cite{Hinton:2006:FLA:1161603.1161605, Smolensky:1986:IPD:104279.104290} or deep Boltzmann Machines \cite{salakhutdinov2009deep}. 

Conditional GANs are used to synthesize images conditioned on different parameters during both training and testing. Examples include conditioning on labels of MNIST to generate digits by Mirza \textit{et al.} \cite{DBLP:journals/corr/MirzaO14}, conditioning on image representations to translate an image between different representations~\cite{pix2pix2017}, and generating panoramic ground-level scenes from aerial images of the same location\cite{zhai2017crossview}. Pathak \textit{et al.} \cite{pathak2016context} generated missing parts in images (i.e., inpainting) using networks trained jointly with adversarial and reconstruction losses and produced sharp and coherent images. Reed \textit{et al.} \cite{pmlr-v48-reed16} synthesized images conditioned on detailed textual descriptions of the objects in the scene, and Zhang \textit{et al.} \cite{han2017stackgan} improved on that by using a two-stage Stacked GAN.

\vspace{-5pt}
\subsection{Cross-Domain Transformations using GANs}
\vspace{-5pt}
Kim \textit{et al.} \cite{pmlr-v70-kim17a} utilized the GAN networks to learn the relation between images in two different domains such that these learned relations can be transferred between the domains. Similar work by Zhu \textit{et al.} \cite{CycleGAN2017} learned mappings between unpaired images using cycle-consistency loss. They assume that a mapping from one domain to the other and back to the first should generate the original image. Both works exploited large unpaired datasets to learn the relation between domains and formulated the mapping task between images in different domains as a generation problem. Zhu \textit{et al.} \cite{CycleGAN2017} compare their generation task with previous works on paired datasets by Isola \textit{et al.} \cite{pix2pix2017}. They conclude that the results with paired images is the upper-bound for their unpaired examples. 


\section{Background on GANs}
\vspace{-5pt}
Generative Adversarial Network architecture \cite{goodfellow2014generative} consists of two adversarial networks: a generator and a discriminator that are trained simultaneously based on the min-max game theory. The generator $G$ is optimized to map a $d$-dimensional noise vector (usually $d$=100) to an image (i.e., synthesizing) that is close to the true data distribution. The discriminator $D$, on the other hand, is optimized to accurately distinguish between the synthesized images coming from the generator and the real images from the true data distribution. The objective function of such a network is 
\begin{equation}
\begin{split}
\resizebox{0.8\hsize}{!}{$%
\stackanchor{min }{G}  \stackanchor{max  }{D} L_{GAN} (G,D) $ = $ E_{x} {\raise.01ex\hbox{$\scriptstyle\sim$}}_{p_{data}(x)} [log D(x) ]+
$%
}%
\\ \hspace*{0.5in}
\resizebox{0.48\hsize}{!}{$%
E_{z}  {\raise.01ex\hbox{$\scriptstyle\sim$}}_{p_z(z)}[log(1 - D(G(z)))],
$%
}%
\end{split}
\end{equation}

\noindent where, $x$ is real data sampled from data distribution ${p_{data}}$ and $z$ is a $d$-dimensional noise vector sampled from a Gaussian distribution ${p_{z}}$. 

Conditional GANs synthesize images looking into some auxiliary variable which may be labels~\cite{DBLP:journals/corr/MirzaO14}, text embeddings~\cite{han2017stackgan,pmlr-v48-reed16} or images~\cite{pix2pix2017,CycleGAN2017,pmlr-v70-kim17a}. In conditional GANs, both the discriminator and the generator networks receive the conditioning variable represented by $c$ in Eqn.~\eqref{eq_cond}. The generator uses this additional information during image synthesis while the discriminator makes its decision by looking at the pair of conditioning variable and the image it receives. Real pair input to the discriminator consists of true image from distribution and its corresponding label while the fake pair consists of synthesized image and the label. For conditional GAN, the objective function is 
\vspace{-10pt}

\begin{equation}\label{eq_cond}
\begin{split}
\resizebox{0.8\hsize}{!}{$%
\stackanchor{min }{G} \stackanchor{max  }{D} L_{cGAN}(G,D) = E_{x,c} {\raise.01ex\hbox{$\scriptstyle\sim$}}_{p_{data}(x, c)} [log D(x,c)] 
$%
}%
\\ \hspace*{0.5in}
\resizebox{0.6\hsize}{!}{$%
+ E_{x', c} {\raise.01ex\hbox{$\scriptstyle\sim$}}_{p_{data}(x',c)}[ log(1 - D(x',c))],
$%
}%
\end{split}
\end{equation}
where $x'$ = $G(z,c)$ is the generated image.

In addition to the GAN loss, previous works (e.g.,~\cite{pix2pix2017,CycleGAN2017,pathak2016context}) have tried to minimize the $L1$ or $L2$ distances between real and generated image pairs. This step aids the generator to synthesize images very similar to the ground truth. Minimizing $L1$ distance generates less blurred images than minimizing the $L2$ distance. That is, using the $L1$ distance increases image sharpness in generation tasks. Therefore, we use the $L1$ distance in our method. The expression to minimize the $L1$ distance is
\vspace{-5pt}
\begin{equation}\label{eq_cond_l1}
\begin{split}
\resizebox{0.7\hsize}{!}{$%
\stackanchor{min  }{G} L_{L1}(G)=E_{x,x'} {\raise.01ex\hbox{$\scriptstyle\sim$}}_{p_{data}(x,x')}[\mid \mid x - x' \mid \mid _1],
$%
}%
\end{split}
\end{equation}

The objective function for such conditional GAN network is the sum of Eqns. \eqref{eq_cond} and~\eqref{eq_cond_l1}.

Considering the synthesis of the ground level imagery $(I_{g})$ from aerial image $(I_{a})$, the conditional GAN loss and $L1$ loss are represented as in Eqns. \eqref{eq_gan} and \eqref{eq_l1}, respectively. For ground to aerial synthesis, the roles of $I_a$ and $I_g$ are reversed.
\vspace{-5pt}
\begin{equation}\label{eq_gan}
\begin{split}
\resizebox{0.8\hsize}{!}{$%
\stackanchor{min }{G} \stackanchor{max  }{D} L_{cGAN}(G,D) = E_{I_{g},I_{a}} {\raise.01ex\hbox{$\scriptstyle\sim$}}_{p_{data}(I_{g},I_{a})} [log D(I_{g},I_{a})]
$%
}%
\\ \hspace*{0.5in}
\resizebox{0.6\hsize}{!}{$%
 + E_{I_a, I_g'} {\raise.01ex\hbox{$\scriptstyle\sim$}}_{p_{data}(I_{a}, I_g')}[ log(1 - D(I_g',I_{a}))],
$%
}%
\end{split}
\end{equation} 
\vspace{-10pt}
\begin{equation}\label{eq_l1}
\begin{split}
\resizebox{0.75\hsize}{!}{$%
 \stackanchor{min  }{G} L_{L1}(G)=E_{I_g,I_g'} {\raise.01ex\hbox{$\scriptstyle\sim$}}_{p_{data}(I_g,I_g')}[\mid \mid I_g - I_g' \mid \mid _1],
$%
}%
\end{split}
\end{equation}
where, $I_g' = G(I_{a})$.
We employ the network of \cite{pix2pix2017} as our baseline architecture. The objective function for the baseline is the sum of conditional GAN loss in Eqn. \eqref{eq_gan} and $L1$ loss in Eqn. \eqref{eq_l1}, as represented in Eqn. \eqref{eq_comb}: 
\vspace{-5pt}
\begin{equation}\label{eq_comb}
\begin{split}
\resizebox{0.65\hsize}{!}{$%
 L_{network} = L_{cGAN} (G,D) + \lambda L_{L1}(G),
$%
}%
\end{split}
\end{equation}
where, $\lambda$ is the balancing factor between the losses.

\begin{figure}[t]
\begin{tabular}{c}
\subcaptionbox{X-Fork architecture. \label{fig:crossview-fork}}{\includegraphics[width=0.80\linewidth]{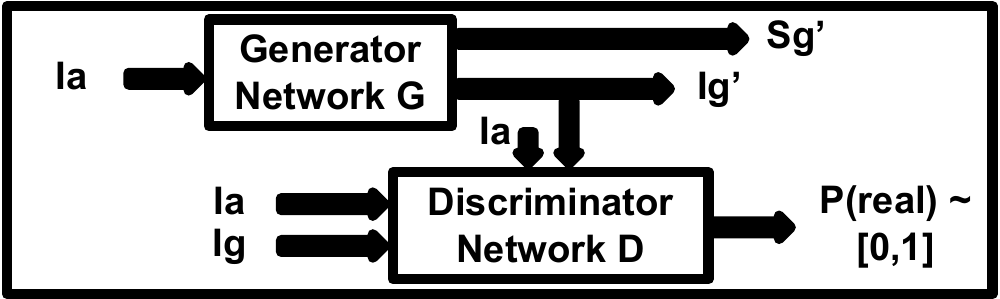}} \\
\subcaptionbox{X-Seq architecture. \label{fig:framework}}{\includegraphics[width=0.94\linewidth]{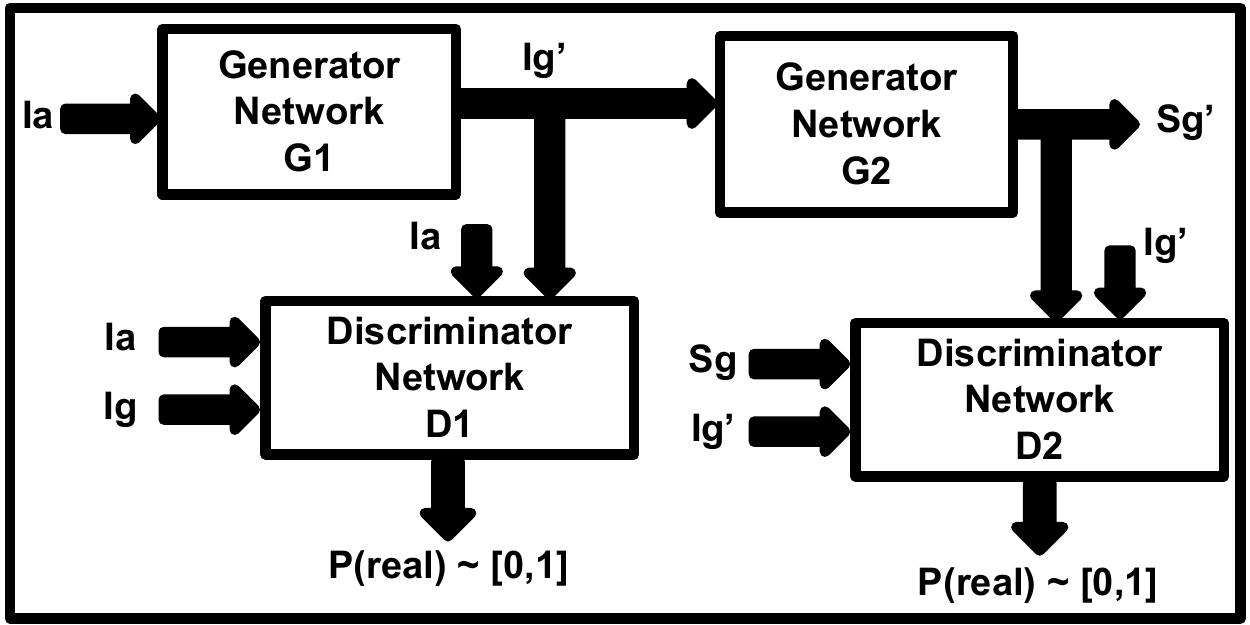}}
\end{tabular}
\vspace{-10pt}
\caption{\small Our proposed network architectures. a) X-Fork: Similar to baseline architecture except that G forks to synthesize image and segmentation map in target view, and b) X-Seq: a sequence of two cGANs, G1 synthesizes target view image that is used by G2 for segmentation map synthesis in corresponding view. In both architectures, $I_a$ and $I_g$ are real images in aerial and ground views, respectively. $S_g$ is the ground-truth segmentation map in street-view obtained using pre-trained RefineNet \cite{Lin:2017:RefineNet}. $I_g'$ and $S_g'$ are synthesized image and segmentation map in ground view.}
 \vspace{-10pt}
\end{figure}

\section{Proposed cGAN-based Approaches}
\vspace{-5pt}
In this section, we propose two architectures for the task of cross-view image synthesis. 

\subsection{Crossview Fork (X-Fork)}
\begin{figure}
\centering
\includegraphics[width=0.48\textwidth]{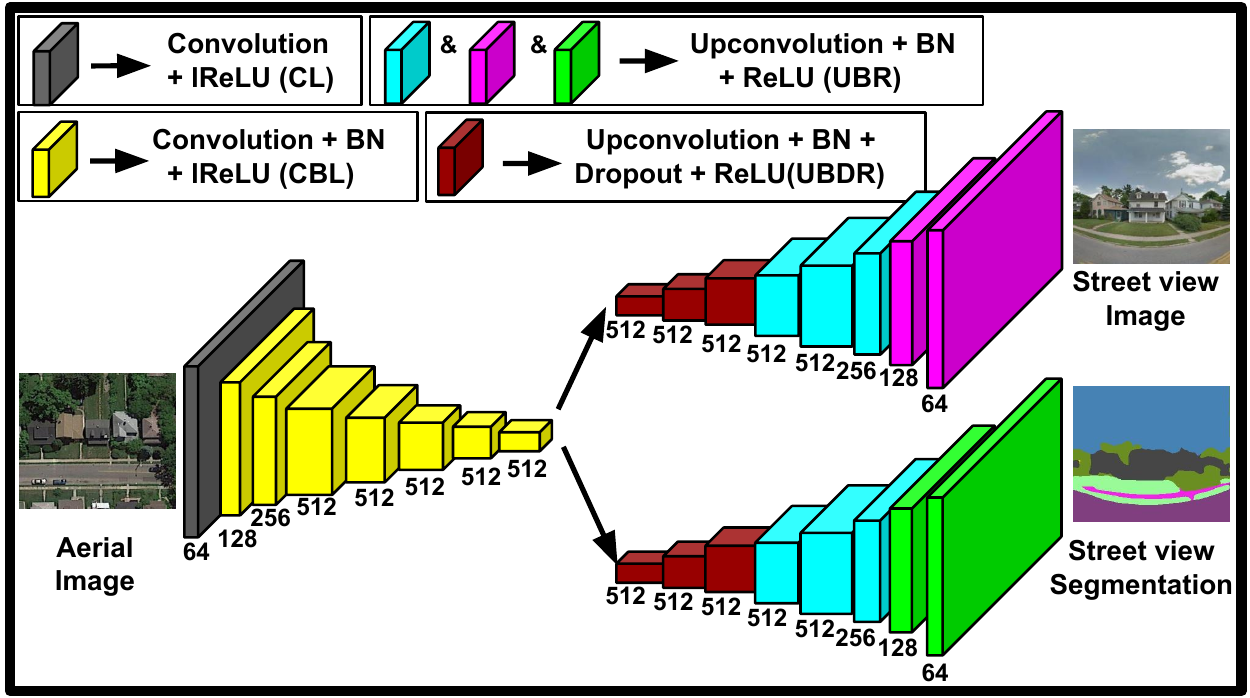}
\vspace{-20pt}
\caption{\small \label{fig:crossview-fork-gen}Generator of X-Fork architecture in Figure \ref{fig:crossview-fork}. BN means batch-normalization layer.
The first six blocks of decoder share weights, forking at the penultimate block. 
The number of channels in each convolution layer are shown below each blocks. }
\vspace{-15pt}
\end{figure}

\vspace{-5pt}
Our first architecture, known as Crossview Fork, is shown in Figure \ref{fig:crossview-fork}. The discriminator architecture is taken from the baseline \cite{pix2pix2017} but the generator network is forked to synthesize images as well as segmentation maps. The fork-generator architecture is shown in Figure \ref{fig:crossview-fork-gen}. The first six blocks of decoder share the weights. This is because the image and segmentation map contain a lot of shared features. The number of kernels used in each layer (block) of the generator are shown below the blocks. 

Even though the X-Fork architecture generates the cross-view image and its segmentation map, the discriminator receives only the real/fake image pairs but not the segmentation pairs during the training. In other words, the generated segmentation map serves as an auxiliary output. Notice that here we are primarily interested in generating higher quality images rather than the segmentation maps. Thus, the conditional GAN loss for this network is still the same as in Eqn. $\eqref{eq_gan}$. To use the segmentation information, in addition to the $L1$ distance between the generated image and the real image, we also include the $L1$ distance between the ground-truth segmentation and the generated segmentation map into the loss. 

\subsection{Crossview Sequential (X-Seq)}
\vspace{-5pt}
Our second architecture uses a sequence of two cGAN networks as shown in Figure \ref{fig:framework}. The first network generates cross-view images similar to the baseline. The second network receives images from the first generator as conditioning input to synthesize the segmentation map in the same view. Thus, the first network is a cross-view cGAN while the second one is an image-to-segmentation cGAN. The whole architecture is trained end-to-end so that both cGANs learn simultaneously. Intuitively, the input-output dependency between the cGANs constrains the generated images and the segmentation maps, and in effect improves the quality of the generated outputs. Training the first network to generate better cross-view images enhances generation of better segmentation maps by the second generator. At the same time, the feedback from the better trained second network forces the first network to improve its generation. Thus, when both networks are trained in tandem, better quality images are generated compared to the baseline.

Replacing $G$ and $D$ in Eqns. \eqref{eq_gan} and \eqref{eq_l1} by $G_1$ and $D_1$, respectively, we obtain the equivalent expressions for losses of cross-view cGAN network in this architecture. For the image-to-segmentation cGAN network, the images generated by $G_1$ are considered as conditioning inputs. We now express the cGAN loss for this network as
\vspace{-5pt}
\begin{equation}\label{eq_gan2}
\begin{split}
\resizebox{0.8\hsize}{!}{$%
\stackanchor{min }{$G_2$} \stackanchor{max  }{$D_2$} L_{cGAN}(G_2,D_2) = E_{I_g',S_g} {\raise.01ex\hbox{$\scriptstyle\sim$}}_{p_{data}(I_g',S_{g})} [log D_2(S_{g},I_g')]+
$%
}%
\\ \hspace*{0.5in}
\resizebox{0.57\hsize}{!}{$%
 E_{S_g',I_g'} {\raise.01ex\hbox{$\scriptstyle\sim$}}_{p_{data}(S_g', I_g')}[ log(1 - D_2(S_g',I_g'))],
$%
}%
\end{split}
\end{equation}
where, $I_g'$ = $G_1(I_a)$ and $S_g'$ = $G_2(I_g')$. 
The $L1$ loss for the image-to-segmentation network is
\vspace{-5pt}
\begin{equation}\label{eq_l1G2}
\begin{split}
\resizebox{0.7\hsize}{!}{$%
 \stackanchor{min  }{$G_2$} L_{L1}(G_2)=E_{S_g,S_g'} {\raise.01ex\hbox{$\scriptstyle\sim$}}_{p_{data}(S_g,S_g')}[\mid \mid S_g - S_g') \mid \mid _1],
$%
}%
\end{split}
\end{equation}

The overall objective function for the X-Seq network is
\vspace{-15pt}

\begin{equation}\label{eq_final}
\begin{split}
\resizebox{0.85\hsize}{!}{$%
L_{X-Seq} = L_{cGAN} (G_1, D_1) + \lambda L_{L1} (G_1) + L_{cGAN} (G_2, D_2) + \lambda L_{L1} (G_2).
$%
}%
\end{split}
\end{equation}

Eqn. \eqref{eq_final} is optimized during the training to learn the parameters $G_1$, $D_1$, $G_2$ and $D_2$.

\section{Experimental Setting}

\subsection{Dataset}
\vspace{-5pt}
For the experiments in this work, we use the cross-view image dataset provided by Vo \textit{et al.} \cite{Vo2016}. This dataset consists of more than one million pairs of street-view and overhead view images collected from 11 different cities in the US. We select 76,048 image pairs from Dayton and create a train/test split of 55,000/21,048 pairs. We call it Dayton Dataset. The images in the original dataset have resolution of 354$\times$354. We resize them to 256$\times$256. Some example images are shown in Figure \ref{fig:obj}.

We also recruit the CVUSA dataset \cite{workman2015wide} for direct comparison of our work with Zhai \textit{et al.} \cite{zhai2017crossview}. This dataset consists of 35,532/8,884 train/test split of image pairs. Following Zhai \textit{et al.}, the aerial images are center-cropped to 224 $\times$ 224 and then resized to 256 $\times$ 256. We only generate a single camera-angle image rather than the panorama. To do so, we take the first quarter of the ground level images and segmentations from the dataset and resize them to 256 $\times$ 256 in our experiments. Please see Figure \ref{fig:cvusa} for some images from the CVUSA dataset.

\begin{figure}
\centering
\includegraphics[width=0.45\textwidth]{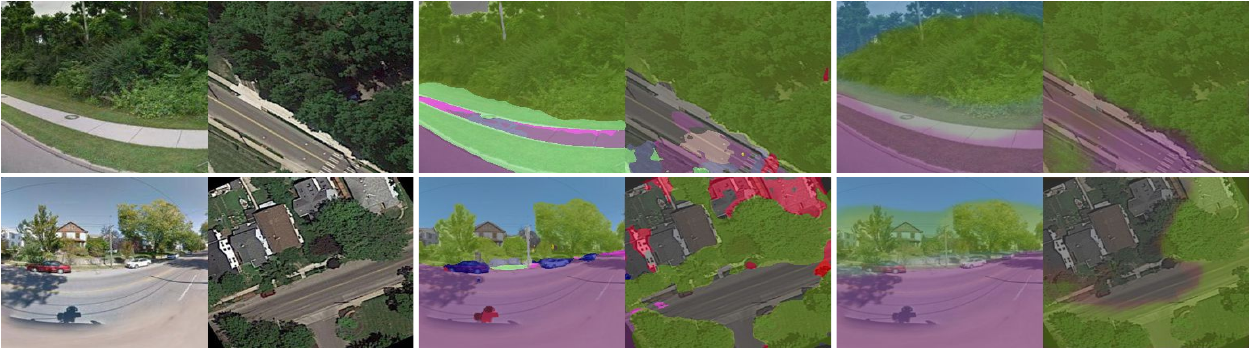}
\vspace{-5pt}
\caption{\small \label{fig:overlay}Original image pairs from training set (left), images with segmentation masks from pre-trained RefineNet \cite{Lin:2017:RefineNet} overlaid on original images (middle) and images with segmentation masks generated by X-Fork network overlaid on original images (right). }
\vspace{-17pt}
\end{figure}

The two networks, X-Fork and X-Seq, learn to generate the target view images and segmentation maps conditioned on source view image. Training procedure requires the images as well as their semantic segmentation maps. The CVUSA dataset has annotated segmentation maps for ground view images, but for Dayton dataset such information is not available. To compensate, we use one of the leading semantic segmentation methods, known as the RefineNet~\cite{Lin:2017:RefineNet}. This network is pre-trained on outdoor scenes of the Cityscapes dataset \cite{Cordts2016Cityscapes} and is used to generate the segmentation maps that are utilized as ground truth maps. These semantic maps have pixel labels from 20 classes (e.g., road, sidewalk, building, vegetation, sky, void, etc). Figure \ref{fig:overlay} shows image pairs from the dataset and their segmentation masks overlaid in both views. As can be seen, the segmentation mask (label) generation process is far from perfect since it is unable to segment parts of buildings, roads, cars, etcetera in images.

\subsection{Implementation Details} 
\vspace{-5pt}
We use the conditional GAN architecture of \cite{pix2pix2017} as the baseline and call it Pix2pix. The generator is an encoder-decoder network with blocks of Convolution, Batch Normalization~\cite{Ioffe:2015:BNA:3045118.3045167} and activation layers. Leaky ReLU with a slope of 0.2 is used as the activation function in the encoder, whereas the decoder has ReLU activation except for its final layer where Tanh is used. The first three blocks of the decoder have a Dropout layer in between Batch normalization and activation layer, with dropout rate of 50\%. The discriminator is similar to the encoder of the generator. The only difference is that the final layer uses sigmoid non-linearity that gives the probability of its input being real.

The used convolutional kernels 
are 4$\times$4 with a stride of 2. The upconvolution in the decoder is Torch\cite{torch} implementation of $SpatialFullConvolution$, and upsamples the input by 2. For the encoder and the discriminator, convolutional operation downsamples the images by 2. No pooling operation is used in the networks. The $\lambda$ used in Eqns. \eqref{eq_comb} and \eqref{eq_final} is the balancing factor between the GAN loss and $L1$ loss. Its value is fixed at 100. Following the idea to smooth the labels by \cite{DBLP:conf/cvpr/SzegedyVISW16} and demonstration of its effectiveness by Salimans \textit{et al.}~\cite{DBLP:conf/nips/SalimansGZCRCC16}, we use one-sided label smoothing to stabilize the training process, replacing 1 by 0.9 for real labels. During the training, we utilized different data augmentation methods like random jitter and horizontal flipping of images. The network is trained end-to-end with weights initialized with a random Gaussian distribution with zero mean and 0.02 standard deviation. It is implemented in Torch \cite{torch}. 


\section{Results} 
\vspace{-5pt}
Our experiments are conducted in \textbf{a2g} (aerial-to-ground) and \textbf{g2a} (ground-to-aerial) directions on Dayton dataset and \textbf{a2g} direction only on CVUSA dataset. We consider image resolutions of 64$\times$64 and 256$\times$256 on Dayton dataset while for experiments on CVUSA dataset, 256$\times$256 resolution images are used. 

First, we run experiments on lower resolution images (64$\times$64) for proof of concept. Encouraging qualitative and quantitative results in this resolution motivated us to apply our methods to higher resolution (256$\times$256) images. The lower resolution experiments are carried out for 100 epochs with batch size of 16, whereas the higher resolution experiments are conducted for 35 epochs with batch size of 4. 

We conduct experiments on CVUSA dataset for comparison with Zhai \textit{et al.}'s work \cite{zhai2017crossview}. Following their setup, we train our architectures for 30 epochs, using the Adam optimizer and moment parameters $\beta 1$ = 0.5 and $\beta 2$ = 0.999. 

It is not straightforward to evaluate the quality of synthesized images~\cite{borji2018pros}. In fact, evaluation of GAN methods continues to be an open problem~\cite{Theis2016a}. A common evaluation method is to show the generated images to human observers and ask their opinion about the images. Human judgment is based on the response to the question: Is this image (image-pair) real or fake? Alternatively, the images generated by different generative models can be pitted against each other and the observer is asked to select the image that looks more real. But in experiments involving natural scenes, such evaluation methods are more challenging as multiple factors often affect the quality of the generated images. For example, the observer may not be sure whether to base his judgment on better visual quality, higher sharpness at object boundaries, or more semantic information present in the image (e.g., multiple objects in the images, more details on objects, etc). Therefore, instead of behavioral experiments, we illustrate qualitative results in Figures \ref{fig:test_64}, \ref{fig:test_256} and \ref{fig:cvusa} and conduct an in-depth quantitative evaluation on test images of two datasets.

\subsection{Qualitative Evaluation}
\begin{figure}
\centering
\includegraphics[width=0.48\textwidth]{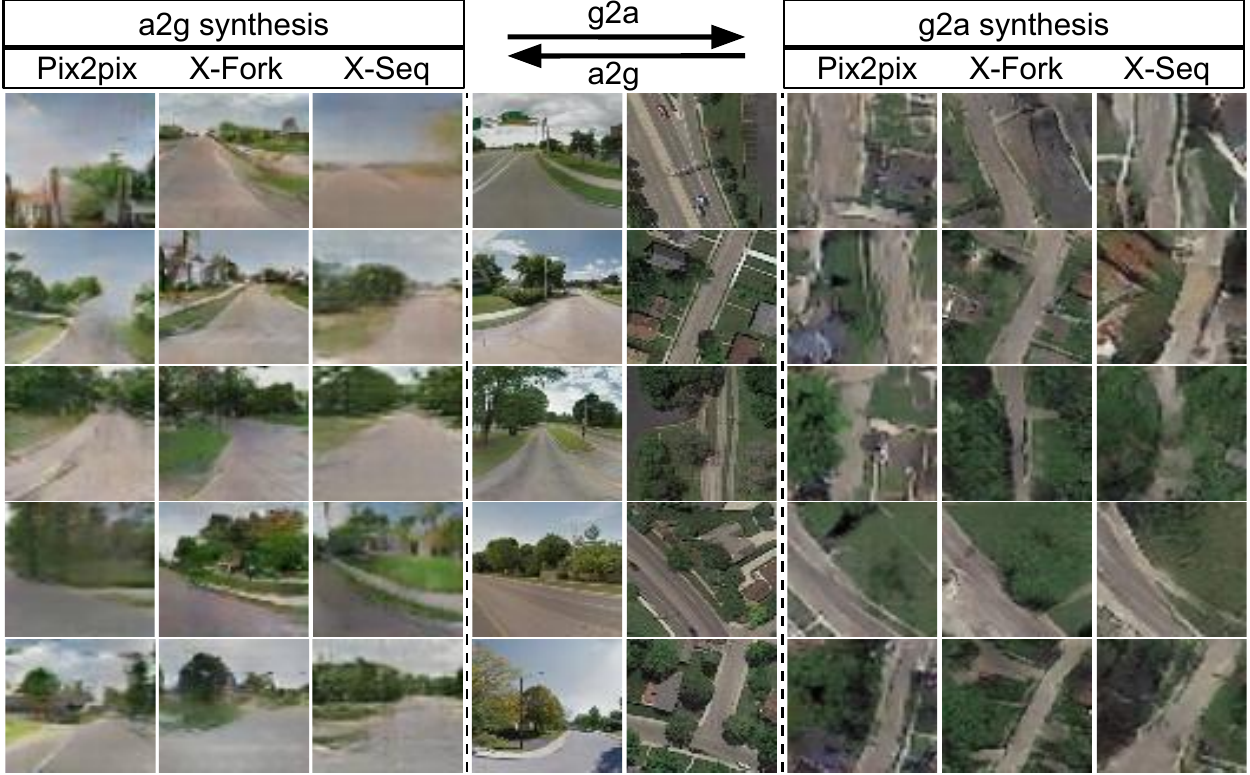}
\vspace{-18pt}
\caption{\small \label{fig:test_64}Example images generated by different methods in lower (64 $ \times$ 64) resolution in  \textbf{a2g} and \textbf{g2a} directions.}
 \vspace{-10pt}
\end{figure}

\vspace{-5pt}
For 64$\times$64 resolution experiments, the networks are modified by removing the last two blocks of CBL from discriminator and encoder, and the first two blocks of UBDR from decoder of the generator. We run experiments on all three methods. Qualitative results are depicted in Figure \ref{fig:test_64}. The results affirm that the networks have learned to transfer the image representations across the views. Generated ground level images clearly show details about road, trees, sky, clouds, and pedestrian lanes. Trees, grass, road, house roofs are well rendered in the synthesized aerial images.

\begin{figure}
\centering
\includegraphics[width=0.48\textwidth]{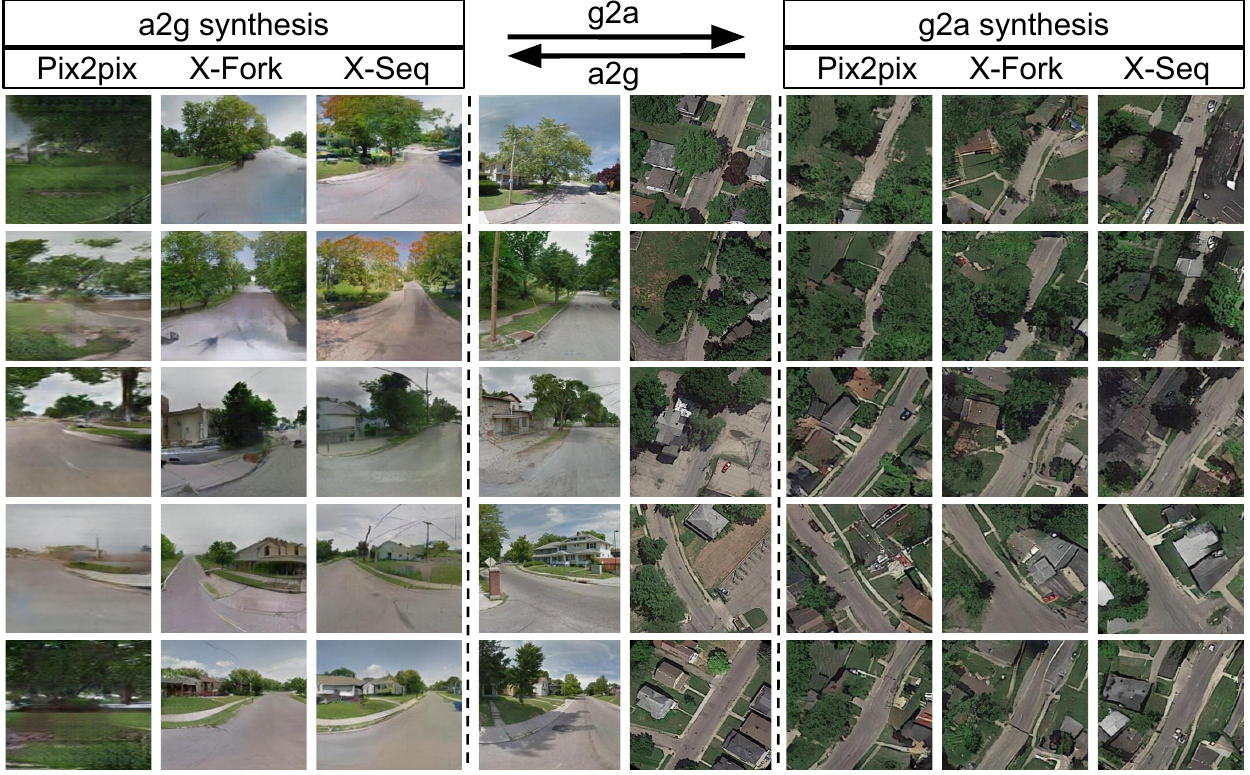}
\vspace{-18pt}
\caption{\small \label{fig:test_256}Example images generated by different methods in higher (256 $\times$ 256) resolution in  \textbf{a2g} and \textbf{g2a} directions.}
\vspace{-12pt}
\end{figure}

\begin{figure}
\centering
\includegraphics[width=0.48\textwidth]{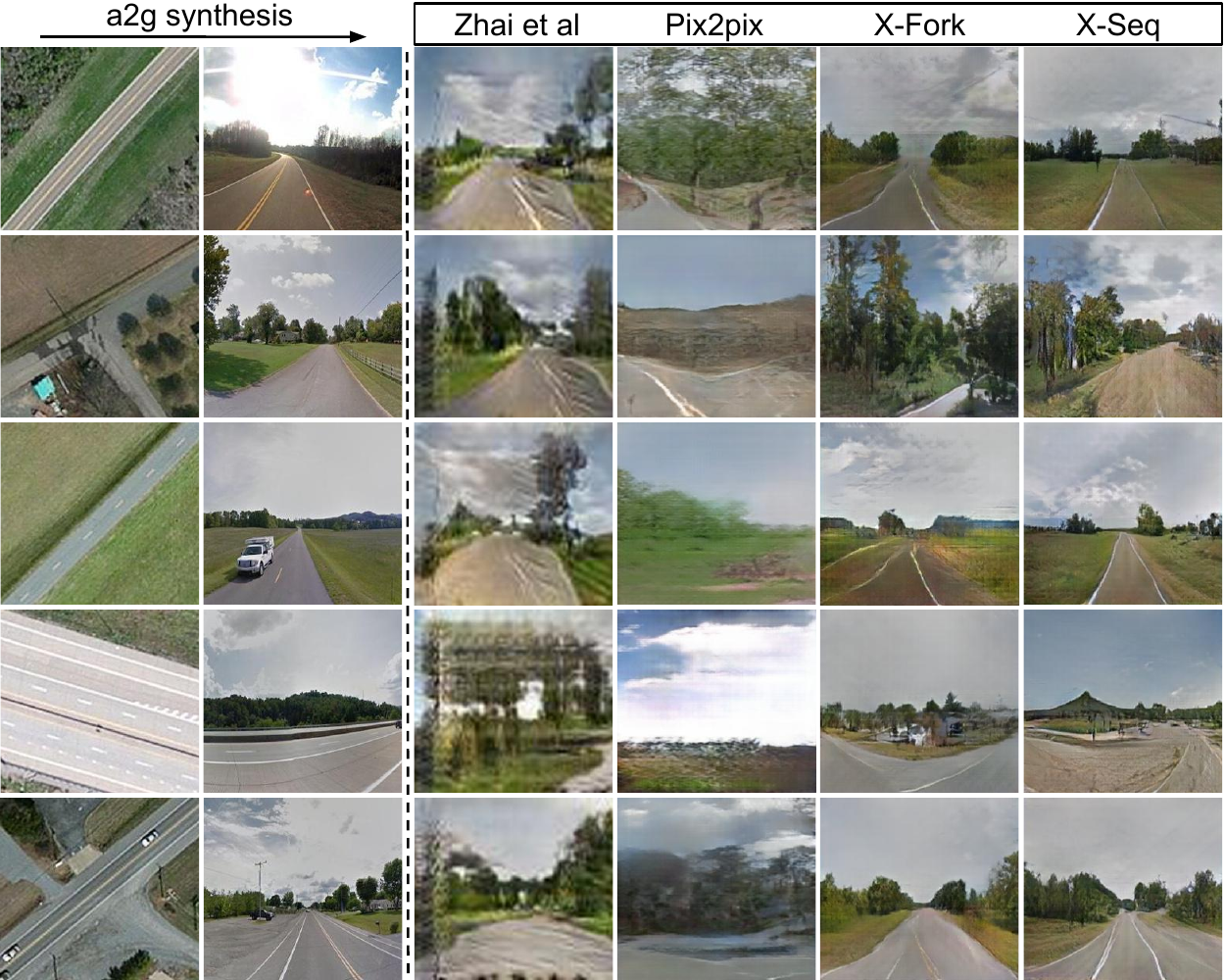}
\vspace{-18pt}
\caption{\small \label{fig:cvusa}Qualitative results of our methods and baselines on CVUSA dataset in \textbf{a2g} direction. First two columns show true image pairs, next four columns show images generated by Zhai \textit{et al.} \cite{zhai2017crossview}, Pix2pix\cite{pix2pix2017}, X-Fork and X-Seq methods, respectively.}
\vspace{-12pt}
\end{figure}

For 256$\times$256 resolution synthesis, we conduct experiments on all three architectures and illustrate the qualitative results on Dayton and CVUSA datasets in Figures \ref{fig:test_256} and \ref{fig:cvusa} respectively. For Dayton dataset, we observe that the images generated in higher resolution contain more details of objects in both views and are less granulated than those in lower resolution. Houses, trees, pedestrian lanes, and roads look more natural. Test results on CVUSA dataset show that images generated by proposed methods are visually better compared to Zhai \textit{et al.} \cite{zhai2017crossview} and Pix2pix \cite{pix2pix2017} methods. 

\subsection{Quantitative Evaluation} 
\vspace{-5pt}
The quantitative results of our experiments on both datasets are presented in Tables \ref{tab:inception_score}-\ref{tab:ssim_psnr_sd}. \textbf{64$\times$64} and \textbf{256$\times$256} in column headers of the tables refer to results obtained for two resolutions of Dayton dataset. Next, we discuss the quantitative measures used to evaluate our methods.
\vspace{-5pt}
\subsubsection{Inception Score}
\vspace{-5pt}
A common quantitative GAN evaluation measure is the \textit{Inception Score}
\cite{DBLP:conf/nips/SalimansGZCRCC16}. 
The core idea behind the inception score is to assess how diverse the generated samples are within a class while being meaningfully representative of the class at the same time.  
\vspace{-5pt}
\begin{equation}
Inception Score, I = exp (E_x D_{KL}(p(y|x)||p(y))),
\end{equation}
where, $x$ is a generated sample and $y$ is its predicted label. 

\begin{table*}[t]
\small
  \renewcommand{\arraystretch}{.9}
  \centering
  \caption{\small KL divergence scores between conditional and marginal probabilities (Inception Score).}
  \vspace{-10pt}
  \label{tab:inception_score}
    \begin{tabular*}{\textwidth}{l @{\extracolsep{\fill}} lccccccccc}
        \toprule
              \multicolumn{1}{l}{\textbf{Dir.}} & \multicolumn{1}{l}{\textbf{Methods}} & \multicolumn{3}{c}{\textbf{64$\times$64}} & \multicolumn{3}{c}{\textbf{256$\times$256}} & \multicolumn{3}{c}{\textbf{CVUSA}}     \\
                \cmidrule(lr){3-5}
\cmidrule(lr){6-8} \cmidrule(lr){9-11}
       $\rightleftarrows$  & & all  & Top-1  & Top-5  & all  & Top-1  & Top-5  & all  & Top-1  & Top-5  \\
         & &  classes &  class &  classes &  classes &  class &  classes &  classes &  class &  classes \\
        \midrule
      &  Zhai \textit{et al.} \cite{zhai2017crossview} & \textbf{--} & \textbf{--} & \textbf{--} & \textbf{--} & \textbf{--} & \textbf{--} & 1.8434 & 1.5171 & 1.8666\\
      &  Pix2pix \cite{pix2pix2017} & 1.8029 & 1.5014 & 1.9300 & 2.8515 & 1.9342 & 2.9083 & 3.2771 & 2.2219 & 3.4312 \\
   a2g &  X-Fork & \textbf{1.9600} & \textbf{1.5908} & \textbf{2.0348} & \textbf{3.0720} & \textbf{2.2402} & \textbf{3.0932} & 3.4432 & 2.5447 & 3.5567\\
   &  X-Seq & 1.8503 & 1.4850 & 1.9623 & 2.7384 & 2.1304 & 2.7674 & \textbf{3.8151} & \textbf{2.6738} & \textbf{4.0077}\\
    \cmidrule(lr){2-11}
   &  Real Data & 2.2096 & 1.6961 & 2.3008 & 3.7090 & 2.5590 & 3.7900 & 4.9971 & 3.4122 & 5.1150\\
     \midrule
  &  Pix2pix \cite{pix2pix2017} & 1.7970 & 1.3029 & 1.6101 & 3.5676 & 2.0325 & 2.8141 & \textbf{--} & \textbf{--} & \textbf{--} \\
  g2a &   X-Fork & \textbf{1.8557} & {1.3162} & \textbf{1.6521} & 3.1342 & 1.8656 & 2.5599 & \textbf{--} & \textbf{--} & \textbf{--}\\
  &   X-Seq & 1.7854 & \textbf{1.3189} & 1.6219 & \textbf{3.5849} & \textbf{2.0489} & \textbf{2.8414} & \textbf{--} & \textbf{--} & \textbf{--}\\
    \cmidrule(lr){2-11}
  &   Real Data & 2.1408 & 1.4302 & 1.8606 & 3.8979 & 2.3146 & 3.1682 & \textbf{--} & \textbf{--} & \textbf{--}\\
        \bottomrule
    \end{tabular*}
\end{table*}

\begin{table*}[t]
\small
  \renewcommand{\arraystretch}{.9}
  \centering
  \caption{\small Accuracies: Top-1 and Top-5.}
  \vspace{-10pt}
  \label{tab:accuracies}
    \begin{tabular*}{\textwidth}{l @{\extracolsep{\fill}} llccccccccccc}
        \toprule
               \multicolumn{1}{l}{\textbf{Dir.}} & \multicolumn{1}{l}{\textbf{Methods}} & \multicolumn{4}{c}{\textbf{64$\times$64}} & \multicolumn{4}{c}{\textbf{256$\times$256}} & \multicolumn{4}{c}{\textbf{CVUSA}}     \\
                \cmidrule(lr){3-6}
\cmidrule(lr){7-10} \cmidrule(lr){11-14}
     $\rightleftarrows$   &  & \multicolumn{2}{c}{Top-1} & \multicolumn{2}{c}{Top-5} & \multicolumn{2}{c}{Top-1} & \multicolumn{2}{c}{Top-5} & \multicolumn{2}{c}{Top-1} & \multicolumn{2}{c}{Top-5} \\
        &  & \multicolumn{2}{c}{Accuracy (\%)} & \multicolumn{2}{c}{Accuracy (\%)} & \multicolumn{2}{c}{Accuracy (\%)} & \multicolumn{2}{c}{Accuracy (\%)} & \multicolumn{2}{c}{Accuracy (\%)} & \multicolumn{2}{c}{Accuracy (\%)} \\
          \midrule
      &  Zhai \textit{et al.} \cite{zhai2017crossview} & \textbf{--} & \textbf{--} & \textbf{--} & \textbf{--} & \textbf{--} & \textbf{--} & \textbf{--} & \textbf{--} & 13.97 & 14.03 & 42.09 & 52.29 \\
      & Pix2pix \cite{pix2pix2017} & 7.90 & 15.33 & 27.61 & 39.07 & 6.8 & 9.15 & 23.55 & 27.00 & 7.33 & 9.25 & 25.81 & 32.67 \\
a2g  &  X-Fork & \textbf{16.63} & \textbf{34.73} & \textbf{46.35} & \textbf{70.01} & 30.00 & 48.68 & 61.57 & 78.84 & \textbf{20.58} & \textbf{31.24} & \textbf{50.51} & \textbf{63.66}\\
  &  X-Seq & 4.83 & 5.56 & 19.55 & 24.96 & \textbf{30.16} & \textbf{49.85} & \textbf{62.59} & \textbf{80.70} & 15.98 & 24.14 & 42.91 & 54.41 \\
    \midrule
    &    Pix2pix \cite{pix2pix2017} & 1.65 & 2.24 & 7.49 & 12.68 & 10.23 & 16.02 & 30.90 & 40.49 & \textbf{--} & \textbf{--} & \textbf{--} & \textbf{--} \\
   g2a & X-Fork & \textbf{4.00} & \textbf{16.41} & \textbf{15.42} & \textbf{35.82} & 10.54 & 15.29 & 30.76 & 37.32 & \textbf{--} & \textbf{--} & \textbf{--} & \textbf{--}\\
  &  X-Seq & 1.55 & 2.99 & 6.27 & 8.96 & \textbf{12.30} & \textbf{19.62} & \textbf{35.95} & \textbf{45.94} & \textbf{--} & \textbf{--} & \textbf{--} & \textbf{--} \\
        \bottomrule
        \vspace{-15pt}
    \end{tabular*}
\end{table*}


We can not use the Inception model because the datasets that we use include natural outdoor images that do not fit into ImageNet classes \cite{imagenet_cvpr09}. To solve this, we use the AlexNet model \cite{DBLP:journals/cacm/KrizhevskySH17} trained on Places dataset \cite{zhou2017places} with 365 categories to compute the inception score. The Places dataset has images similar to those in our datasets. The scores are reported in Table \ref{tab:inception_score}. The scores for X-Fork generated images are closest to that of real data distribution for Dayton dataset in lower resolution in both directions and also in higher resolution in \textbf{a2g} direction. The X-Seq method works best for CVUSA dataset and for \textbf{g2a} synthesis in higher resolution over Dayton dataset.

We observe that the confidence scores predicted by the pre-trained model on our dataset are dispersed between classes for many samples and not all the categories are represented by the images. Therefore, we compute inception scores on Top-1 and Top-5 classes, where "Top-k" means that top k predictions for each image are unchanged while the remaining predictions are smoothed by an epsilon equal to (1 - $\sum$(top-k predictions))/(n-k classes). Results on top-k classes follow a similar pattern as in all classes (except for Top-1 class on lower resolution in \textbf{g2a} over Dayton dataset).

In addition to inception score, we compute the top-k prediction accuracy between real and generated images. We use the same pre-trained Alexnet model to obtain annotations for real images and class predictions for generated images. We compute top-1 and top-5 accuracies. Results are shown in Table \ref{tab:accuracies}. For each setting, accuracies are computed in two ways: 1) considering all images, and 2) considering real images whose top-1 (highest) prediction is greater than 0.5. Below each accuracy heading, the first column considers all images whereas the second column computes accuracies the second way. For lower resolution images on Dayton dataset and for experiments on CVUSA dataset, X-Fork method outperforms the remaining methods. For higher resolution images, our methods show dramatic improvements over Pix2pix in the \textbf{a2g} direction, whereas X-Seq works best in the \textbf{g2a} direction. 

\begin{table}[t]
 \small
  \centering
  \renewcommand{\arraystretch}{.8}
  \renewcommand{\tabcolsep}{.75mm}  
  
  \caption{\small KL Divergence between model and data distributions.}
  \vspace{-10pt}
  \label{tab:model_data_KL}
    \begin{tabular}{llccc}
        \toprule
       \textbf{Dir.} & \textbf{Method} & \textbf{64$\times$64} & \textbf{256$\times$256} & \textbf{CVUSA}\\
        \midrule
        & Zhai \textit{et al.} \cite{zhai2017crossview} & \textbf{--} & \textbf{--} & $ 27.43 \pm 1.63 $\\
        & Pix2pix \cite{pix2pix2017} & $ 6.29 \pm  0.8 $ & $ 38.26 \pm 1.88 $ &   $ 59.81 \pm 2.12 $\\
    a2g & X-Fork &   $ \textbf{3.42} \pm \textbf{0.72} $ & $6.00 \pm 1.28$ &  $ \textbf{11.71} \pm \textbf{1.55} $
\\
    & X-Seq & $ 6.22 \pm 0.87 $ & $ \textbf{5.93} \pm  \textbf{1.32}$ & $ {15.52} \pm {1.73} $\\
    \midrule 
    & Pix2pix \cite{pix2pix2017} & $ 6.39 \pm  0.90 $ & $ 7.88 \pm 1.24 $ &   \textbf{--} \\
    g2a & X-Fork &   $ \textbf{4.45} \pm \textbf{0.84} $ & $\textbf{6.92} \pm \textbf{1.15} $ & \textbf{--}
\\
    & X-Seq & $ 7.20 \pm 0.92 $ & $ {7.07} \pm  {1.19}$ & \textbf{--}\\
        \bottomrule
        \vspace{-25pt}
    \end{tabular}
\end{table}

\begin{table*}[htbp]
 \small
   \renewcommand{\arraystretch}{.8}
  \centering
  \vspace{-15pt}
  \caption{\small SSIM, PSNR and Sharpness Difference between real data and samples generated using different methods.}
  \vspace{-10pt}
  \label{tab:ssim_psnr_sd}
  
    \begin{tabular*}{\textwidth}{l @{\extracolsep{\fill}} lcccccccccc}
        \toprule
             \multicolumn{1}{l}{\textbf{Dir.}} &   \multicolumn{1}{l}{\textbf{Methods}} & \multicolumn{3}{c}{\textbf{64$\times$64}} & \multicolumn{3}{c}{\textbf{256$\times$256}} & \multicolumn{3}{c}{\textbf{CVUSA}}     \\
                \cmidrule(lr){3-5}
\cmidrule(lr){6-8} \cmidrule(lr){9-11}
    $\rightleftarrows$     &  & SSIM & PSNR & Sharp Diff & SSIM & PSNR & Sharp Diff & SSIM & PSNR & Sharp Diff \\
        \midrule
   & Zhai \textit{et al.} \cite{zhai2017crossview} & \textbf{--} & \textbf{--} & \textbf{--} & \textbf{--} & \textbf{--} & \textbf{--}  & 0.4147 & 17.4886 & 16.6184 \\
    & Pix2pix \cite{pix2pix2017} & 0.4808 & 19.4919 & 16.4489 & 0.4180 & 17.6291 & 19.2821  & 0.3923 & 17.6578 & 18.5239 \\
  a2g & X-Fork & 0.4921 & 19.6273 & 16.4928 & 0.4963 & 19.8928 & 19.4533  & \textbf{0.4356} & \textbf{19.0509} & \textbf{18.6706}\\
   & X-Seq & \textbf{0.5171} & \textbf{20.1049} & \textbf{16.6836} & \textbf{0.5031} & \textbf{20.2803} & \textbf{19.5258}  & 0.4231 & 18.8067 & 18.4378 \\
    \midrule
  &  Pix2pix \cite{pix2pix2017} & 0.3675 & 20.5135 & {14.7813} & 0.2693 & 20.2177 & 16.9477 & \textbf{--} & \textbf{--} & \textbf{--} \\
  g2a &  X-Fork & \textbf{0.3682} & \textbf{20.6933} & \textbf{14.7984} & \textbf{0.2763} & \textbf{20.5978} & \textbf{16.9962}  &  \textbf{--} & \textbf{--} & \textbf{--}\\
 &   X-Seq & {0.3663} & {20.4239} & {14.7657} & {0.2725} & {20.2925} & {16.9285}  & \textbf{--} & \textbf{--} & \textbf{--} \\
        \bottomrule
        \vspace{-20pt}
    \end{tabular*}
\end{table*}  

\vspace{-5pt}

\subsubsection{KL(model $\|$ data)}
\vspace{-5pt}
We next compute the KL divergence between the model generated images and the real data distribution for quantitative analysis of our work, similar to some generative works~\cite{che-2016-reg-gan,tu_etal_nips17_d2gan}. We again use the same pre-trained Alexnet as in the previous subsection. The lower KL score implies that the generated samples are closer to the real data distribution. The scores are provided in Table \ref{tab:model_data_KL}. As it can be seen, our proposed methods generate much better results than existing generative methods on both datasets. X-Fork generates images very similar to real distribution in all experiments except on the higher resolution \textbf{a2g} experiment where X-Seq is slightly better than X-Fork. 

\subsubsection{SSIM, PSNR and Sharpness Difference}
\vspace{-5pt}
As in some generative works~\cite{journals/corr/MathieuCL15,DBLP:conf/cvpr/LedigTHCCAATTWS17,DBLP:conf/cvpr/ShiCHTABRW16,DBLP:journals/corr/ParkYYCB17}, we also employ Structural-Similarity (SSIM), Peak Signal-to-Noise Ratio (PSNR) and Sharpness Difference measures to evaluate our methods. 

SSIM measures the similarity between the images based on their luminance, contrast and structural aspects. SSIM value ranges between -1 and +1. A higher value means greater similarity between the images being compared. It is computed as
\vspace{-18pt}

\begin{equation}
\resizebox{0.7\hsize}{!}{$
SSIM (I_g, I_g')   = \dfrac{(2 \mu_{I_g} \mu_{I_g'} + c_1)(2 \sigma_{{I_g}{I_g'}}+c_2)}{(\mu_{I_g}^2 + \mu_{I_g'}^2 + c_1)(\sigma_{I_g}^2 + \sigma_{I_g'}^2+c_2)}
$
}\vspace{-5pt}
\end{equation}

PSNR measures the peak signal-to-noise ratio between two images to assess the quality of a transformed (generated) image compared to its original version. The higher the PSNR, the better is the quality of generated image. It is computed as
\vspace{-8pt}
\begin{equation}
\resizebox{0.5\hsize}{!}{$
PSNR (I_g, I_g')   = 10 log_{10} ( \dfrac{max^2_{I_g'}}{mse})
$
}
\end{equation}
\noindent where, 
$
mse (I_g, I_g')   = \dfrac{1}{n} \sum_{i = 1}^{n} (I_g[i] - I_g'[i])^2 
,$

\noindent and $max_{I_{g'}}$ = 255 (maximum pixel intensity value).

Sharpness difference measures the loss of sharpness during image generation. To compute the sharpness difference between the generated image and the true image, we follow~\cite{journals/corr/MathieuCL15} and compute the difference of gradients between the images as
\vspace{-10pt}
\begin{equation} \label{eq_sharpdiff}
\resizebox{0.7\hsize}{!}{$
SharpDiff.(I_g, I_g')   = 10 log_{10} ( \dfrac{max^2_{I_g'}}{grads}),
$
}
\end{equation}
\noindent where,
$grads = $$\dfrac{1}{N}\sum_{i} \sum_{j}$$ (|(\triangledown_{i}I_g + \triangledown_{j}I_g) $-$ (\triangledown_{i}I_g' + \triangledown_{j}I_g')|)$

\noindent and, 
$
\triangledown_{i}I = |I_{i,j} - I_{i-1,j}| 
$
, 
$
\triangledown_{j}I = |I_{i,j} - I_{i,j-1}|.
$


Sharpness difference in Eqn.~\eqref{eq_sharpdiff} is inverse of $grads$. We would like the $grads$ to be small, so the higher the overall score the better. 

The scores are reported in Table \ref{tab:ssim_psnr_sd}. Over Dayton dataset, X-Seq model works the best in \textbf{a2g} direction while X-Fork outperforms the rest in the \textbf{g2a} direction. On CVUSA, X-Fork improves over Zhai \textit{et al.} by 5.03\% in SSIM, 8.93\% in PSNR, and 12.35\% in Sharpness difference. 

Because there is no consensus in evaluation of GANs, we had to use several scores. Theis \textit{et al.} \cite{Theis2016a} show that these scores often do not agree with each other and this was observed in our evaluations as well. So, it is difficult to infer whether X-Fork or X-Seq is better. We find that the proposed methods are consistently superior to the baselines.

\subsection{Generated Segmentation Maps} 
\vspace{-5pt}
Our methods generate semantic segmentation maps along with the real images in cross-view. The overlay of segmentation maps generated by X-Fork network (pre-trained RefineNet) on Dayton images are presented in the last (second) column of Figure \ref{fig:overlay}. Please see supplementary materials for more qualitative results on semantic segmentation. The overlaid images show that the network is able to learn the semantic representations of object classes. For quantitative analysis, segmentation maps generated by our methods are compared against the segmentation maps obtained by applying RefineNet~\cite{Lin:2017:RefineNet} to the target images. We compute per-class accuracies and mean IOU for the most common classes in our datasets: `vegetation', `road' and `building' in aerial segmentation maps plus the `sky' in ground segmentations. The scores are reported in Table \ref{tab:segmap_evaluations}. Even though X-Fork does better than X-Seq, we find that both methods achieve good scores for segmentation.

\begin{table}[t]
 \small
  \centering
  \renewcommand{\arraystretch}{.9}
  \renewcommand{\tabcolsep}{.75mm}  
 \caption{\small Evaluation Scores for segmentation maps.}
 \vspace{-10pt}
  \label{tab:segmap_evaluations}
    \begin{tabular}{lcccc}
        \toprule
               \multicolumn{1}{l}{\textbf{Methods}} & \multicolumn{2}{c}{\textbf{a2g}} & \multicolumn{2}{c}{\textbf{g2a}}   \\
                \cmidrule(lr){2-3}
\cmidrule(lr){4-5} 
     \textbf{} & {\textbf{Per-class acc.}} & \textbf{mIOU} & {\textbf{Per-class acc.}} & \textbf{mIOU} \\
         \midrule
     X-Fork & \textbf{0.6262} & \textbf{0.4163} & \textbf{0.5473} & \textbf{0.2157} \\
     X-Seq & {0.4783} & {0.3187} & {0.4990} & {0.2139} \\
        \bottomrule
    \end{tabular}
\end{table}

\subsection{$k$NN} 
\begin{figure}
\centering
\includegraphics[width=0.48\textwidth]{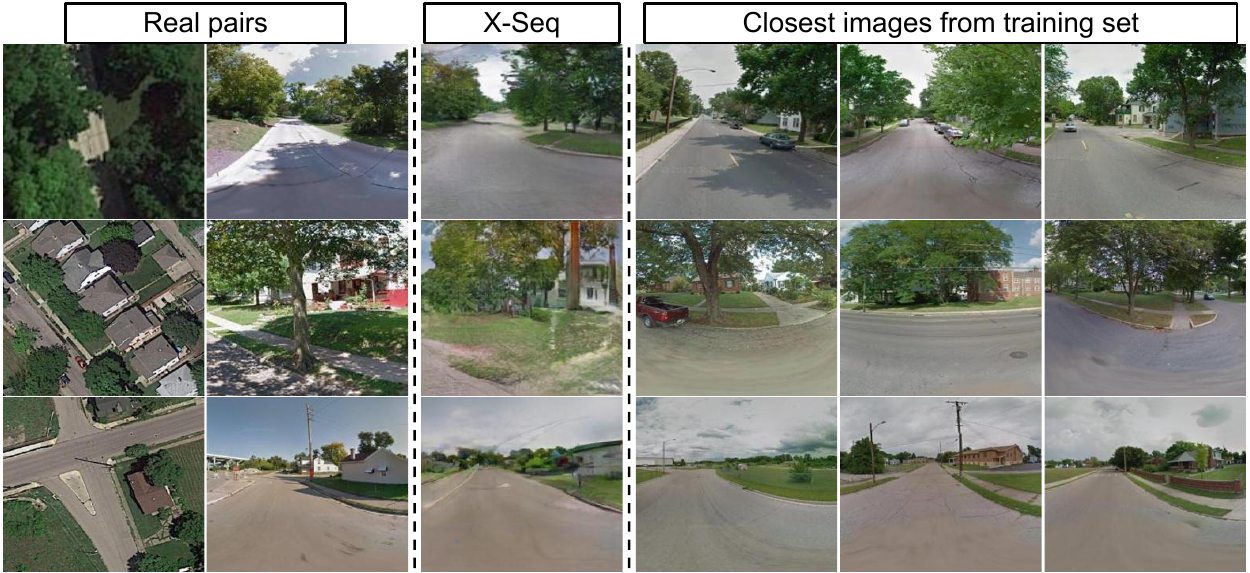}
\vspace{-20pt}
\caption{\small \label{fig:knn}Along the columns, we show real image pairs, corresponding street-view image synthesized by X-Seq method and three nearest images in the training set retrieved by computing $L1$ distance between generated image and training set images.}
\vspace{-15pt}
\end{figure}

\vspace{-5pt}
Here, we test whether our proposed architectures have actually learned the representations between images in two views rather than just memorizing the blocks from training images to generate new ones. For this, we pick three images from the training set that are closest to the generated images in terms of $L1$ distance. As shown in Figure \ref{fig:knn}, the generated images have subtle differences with the training set images implying that our network has indeed learned important semantic representations in input view needed to transform the source image to target view. 

\vspace{-5pt}
\section{Discussion and Conclusion}  
\vspace{-5pt}
We explored image generation using conditional GANs between two drastically different views. Generating semantic segmentations together with images in target view helps the networks learn better images compared to the baselines. Extensive qualitative and quantitative evaluations testify the effectiveness of our methods. Using higher resolution images provided significant improvements in visual quality and added more details to synthesized images. The challenging nature of the problem leaves room for further improvements. Code and data will be shared.

\noindent \textbf{Acknowledgement.} We would like to thank NVIDIA for the donation of the Titan-X GPUs used in this work.

\newpage

{\small
\bibliographystyle{ieee}
\bibliography{sample}
}
\newpage




\end{document}